\documentclass[letterpaper]{article} 
\usepackage{aaai23}  
\usepackage{times}  
\usepackage{helvet}  
\usepackage{courier}  
\usepackage[hyphens]{url}  
\usepackage{graphicx} 
\usepackage{natbib}  
\usepackage{caption} 
\usepackage{algorithm}
\usepackage{algorithmic}

\usepackage{newfloat}
\usepackage{listings}
\DeclareCaptionStyle{ruled}{labelfont=normalfont,labelsep=colon,strut=off} 
\floatstyle{ruled}
\newfloat{listing}{tb}{lst}{}
\floatname{listing}{Listing}
\graphicspath{{figs}}

\usepackage{tipa}
\usepackage{booktabs}

\title{NoLoR: An ASR-Based Framework for Expedited\\Endangered Language Documentation with\\Neo-Aramaic as a Case Study}
\author{
    Matthew Nazari \textsuperscript{\rm 1,\rm 2}
}
\affiliations{
    \textsuperscript{\rm 1} John A. Paulson School of Engineering and Applied Sciences, Harvard University\\
    \textsuperscript{\rm 2} The North-Eastern Neo-Aramaic Database Project, University of Cambridge \\

    matthewnazari@college.harvard.edu\\

}

\usepackage{bibentry}

\usepackage{xcolor}

\begin{document}

\maketitle

\begin{abstract}
    The documentation of the Neo-Aramaic dialects before their extinction has been described as the most urgent task in all of Semitology today. The death of this language will be an unfathomable loss to the descendents of the indigenous speakers of Aramaic, now predominantly diasporic after forced displacement due to violence. This paper develops an ASR model to expedite the documentation of this endangered language and generalizes the strategy in a new framework we call NoLoR.
\end{abstract}

\section{Introduction}
\begin{figure}[t]
\centering
\includegraphics[width=\columnwidth]{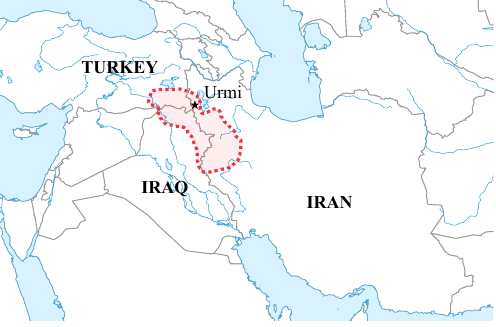}
\caption{Area where North-Eastern Neo-Aramaic dialects are spoken by indigenous communities (in red). The city of Urmi is the center for one of the most common dialects.}
\label{fig:map}
\end{figure}

Aramaic is the oldest continuously spoken Semitic language in the world. Today, the language is highly endangered and is expected to be within its final generation or two of fluent speakers. Speakers of the Neo-Aramaic dialects, which are comprised of Christian and Jewish communities across the Middle East (see Fig. \ref{fig:map}), have been the victims of much violence over the past century, displacing them from their homelands. Such violence includes the Assyrian genocide of World War I \cite{gaunt_2020} and the systematic ethnic cleansing by Islamic State terrorist groups \cite{uscirf}. The death of this language will be an unfathomable loss for the descendents of these communities, to whom language is an essential part of their culture and identity.

This problem is not unique to Aramaic. Over 40\% of spoken languages are endangered, and in the next 100 years, about 90\% of spoken languages are expected to become extinct \cite{moseley_nicolas_2010}. Unfortunately, technologies like the internet and smartphones only further marginalize endangered language by disproportionally promoting the usage of high-resource languages like English \cite{nllb}. Although the field of low-resource NLP appears to be addressing this disparity, it is predominantly focused on developing luxury technologoies, such as translators for already documented languages. However, such work overlooks languages that are underdocumented, without a written tradition, or with very few speakers remaining. Attention must be paid to such situations where the priority is to document the language before it is extinct.

\subsection{Endangered Language Documentation}

In order to preserve endangered languages, they must be documented. The field of endangered language documentation emerged as its own field of linguistics in the 1990s. In this field, researchers document the grammar and oral history of a language by collecting samples of speech accompanied with a written transcription. However, there is a ``transcription bottleneck'' limiting the scale of documentation efforts \cite{jiatong}. Annotating speech data is an extremely costly, lengthy, and esoteric process only performable by the few experts of that language. This severely limits the speed and capacity at which documentation efforts can operate. Hence, it is important that tools and frameworks exist to address this problem.

This paper constructs a framework for documenting endangered languages, employing ASR to overcome the transcription bottleneck. The effectiveness of this strategy is demonstrated by the application of NoLoR in the documentation of the Neo-Aramaic dialect of the Assyrian Christians of Urmi (C. Urmi).

\subsection{The NoLoR Framework}

\begin{figure}[t]
    \centering
    \includegraphics[width=\columnwidth]{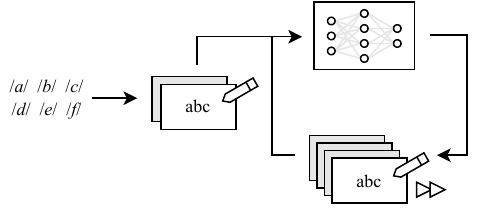}
    \caption{The NoLoR framework describes a positive feedback loop following two preliminary stages. In this loop, language documentation teams collect more data and their efficiency to transcribe increases.}
    \label{cycle} 
\end{figure}

The No- to Low-Resource (NoLoR) framework outlines the strategy of developing an ASR model to expedite the documentation of C. Urmi in a manner generalizable to other language documentation efforts. The NoLoR framework is comprised of four parts:

\paragraph{Step 1: Defining a Phonemic Orthography.} Many endangered languages are unwritten or may not have any pre-existing documentation efforts. Defining a phonemic orthography is hence necessary to begin the documentation process or align with pre-existing ones.

\paragraph{Step 2: Building an Initial Dataset.} After collecting and transcribing interviews, folktales, and other narrations, we can construct an initial dataset ready for a computational pipeline.

\paragraph{Step 3: Training an ASR Model.} An ASR model can be trained from the initial dataset by fine-tuning a pretrained ASR model using careful, task-specific optimizations.

\paragraph{Step 4: Expanding the Dataset.} The transcription of subsequently collected data will be expedited and more data will then be created.

Steps 1 and 2 allows us to enter the positive feedback loop created by steps 3 and 4. Each iteration of this loop produces an ASR model that can become accurate enough to transcribe samples with only minimal expert intervention.

\subsection{Contributions}

This paper proves the NoLoR framework to have been an impactful contribution to both the Assyrian community and to the field of endangered language documentation through its application to the documentation of C. Urmi. Furthermore, this paper contributes a speech dataset, an ASR model, and AssyrianVoices, an online application to crowdsource speech data specifically for future machine learning tasks.

\section{Related Work}

\subsection{Within Language Documentation}
Researchers regularly use software to assist in documenting languages. Such software includes FieldWorks Language Explorer, which allows users to define a lexicon of the target language and to morphologically tag text \cite{flex}. The software SayMore provides a means of file organization and supplies an interface for audio transcription \cite{saymore}. Softwares like these are quite outdated, and they mainly aim to improve transcribing efficiency with convinient interfaces. Meanwhile, NoLoR is itself a framework that aims to automate the transcription process entirely.

\subsection{Within Low-Resource NLP}

The low-resource NLP community has frequently used endangered languages as a case study in various tasks, but not often ASR \cite{hedderich,ranathunga,adams-etal-2017-cross}. A main goal of this field is to eliminate communication boundaries between communities. Almost no work has explicitly sought to uplift the status of endangered languages by improving the workflow of fieldworkers documenting them.

The exception to this is one work that offers a general-purpose ASR model to assist in the documentation of the endangered Muyu language of New Guinea \cite{zahrer-etal-2020-towards}. This ASR model was trained to learn a universal phonetic representation of human speech by training on speech data from 3 languages (American English, Austrian German, and Slovenian). Such model can then be used on other lanuages, including endangered ones. To overcome the low-resource barrier, we also exploit a general representation of human speech by fine-tuning a wav2vec 2.0 pretrained on not 3 languages, but 53 languages then a phonetically adjacent one. This difference, alongside many others, results in the ASR model produced by NoLoR to vastly outperform with much less data (see Table \ref{table:compare}). Additionally, NoLoR rewards researchers by improving the accuracy of the ASR as more documentation work is collected through periodic fine-tuning the ASR. Finally, this paper emperically demonstrates that the ASR model was able to meaningfully improve transcription times.

\begin{table}[t]
    \centering
    \resizebox*{\columnwidth}{!}{
    \begin{tabular}{ccccc}\toprule
         & Language & Labelled & Unlabelled & CER (\%) \\ \midrule
        \citealt{lrspeech}  & Lithuanian & 1.35hr & 11.8hr & \textbf{10.3} \\
        Zahrer et al. \citeyear{zahrer-etal-2020-towards}  & Muyu & 2.05hr & \phantom{0}0.0hr & 48.3 \\
        This paper & Neo-Aramaic & \textbf{0.58hr} & \textbf{\phantom{0}0.0hr} & 12.5 \\ \bottomrule
    \end{tabular}
    }
    \caption{Comparison of previous examples of low-resource ASR and ours}
    \label{table:compare}
\end{table}

Examining the literature more generally, there have been many attempts to train an ASR model with low training data. However, even works that fall under the category of ``extremely low-resource'' assume dozens of hours of unlabelled data \cite{lrspeech}. Our approach, using a phonemic orthography performing careful optimizations, proves to be on par with such work with far less data.

We also develop AssyrianVoices, an online application for crowdsourcing both speech samples of existing text examples and new text samples. This format is directly inspired by the Common Voice project started by Mozilla \cite{commonvoice}. A seperate platform was developed for Neo-Aramaic in order to allow for the display of text in multiple transliteration schemes to maximize accessibility, something not possible with Common Voice.

\section{Specifying a Phonemic Orthography}

Specifying a phonemic orthography allows us to:
\begin{enumerate}
    \item[a)] work with existing language documentation efforts or pave the way for future ones
    \item[b)] minimize the amount of initial data we need to collect to train a minimally viable ASR model
\end{enumerate}
In the case of C. Urmi, we align ourselves with existing work and use the orthographic standard set by Geoffrey Khan in his description of the dialect \cite{khan2016neo}.

\subsection{The Importance of Orthography}

In language documentation, the language of interest may not have a written tradition, or it may use a seperate esoteric literary language that is greatly misaligned with the vernacular. This was especially true for langauges of indigenous communities, such as the Navajo language of North America and the Neo-Aramaic dialects. In these cases, a very first step in documentation is to design an effective orthography \cite{cahill2008factors}. A phonemic orthography is one that assigns each unique unit sound in a language (phone) to a unique character. This scheme has been classically accepted by linguists as most ideal \cite{pike1971phonemics}. However, most contemperary opinions acknowledge the importance of other features of a language beyond phonemic analysis that ought to be represented in the orthography \cite{venezky2004search}. In English, whose orthography is not phonemic, apostraphes to represent possessive constructs are an example of this. In the orthography we selected for C. Urmi, the double hyphen `{\small=}' indicates an enclitic boundary.

\begin{table}[t]
    \centering
    \begin{tabular}{l@{\hspace{2em}}l@{\hspace{2em}}l}\toprule
        Natural & Phonemic & Phonetic \\ \midrule
        `strewn' & \textipa{/strun/} & \textipa{[st\*ru:n]} \\
        `tenth' & \textipa{/tEnT/} & \textipa{[t\super h\~E\r*nT]} \\
        `zine' & \textipa{/zin/} & \textipa{[zi:n]} \\
        `clean' & \textipa{/klin/} & \textipa{[k\textsuperimposetilde{l}i:n]} \\ \bottomrule
    \end{tabular}
    \caption{A phonemic orthography is a sweet-spot representation of human language from both a language documentation perspective and a machine learning one.}
\end{table}

\begin{figure}[H]
    \centering
    \includegraphics[width=\columnwidth]{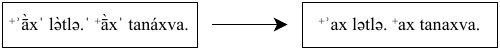}
    \caption{Example of refining original transcriptions for machine learning tasks.}
    \label{fig:transcription}
\end{figure}

Unsurprisingly, a phonemic orthography is a great written representation of spoken language in machine learning tasks since the written and the waveform representations of human speech are in principle being maximally aligned \cite{9414478, khare2021low}. This is different from ``natural'' orthographies, like that of standard French or English, which have frequent irregularities and letters with multiple possible sounds, making it more difficult for speech recognition models to learn. One might then consider a phonetic orthography, like the International Phonetic Alphabet (IPA), but this is demonstrated as being too nuanced for ASR models to learn without massive amounts of examples.

The orthography of C. Urmi is phonemic with special symbols for certain features like enlitic boundaries and suprasegmental emphasis. Most of these are left in, with the exception of markers for features notorious for being difficult to pick up on such as vowel stress and intonation boundaries \cite{bosch2000emotions,vicsi2006prosodic}.

\section{Building an Initial Dataset}

Building an initial dataset from preliminary documentation efforts allows us to train an ASR model that can expedite the transcription of future speech samples. In the case of C. Urmi, a labelled speech dataset of 35 minutes was built. This dataset has been made publically available\footnote{\url{https://huggingface.co/datasets/mnazari/urmi-assyrian-voice}} and is released under a Creative Commons CC0 license to encourage follow-up work.

\subsection{Optimizing Data for Speech Recognition}

Preliminary documentation efforts will produce a dataset $\mathcal{D} = \{(x_i, y_i)\}$ of audio-transcript pairs. Since a priority in preserving oral history is eliciting speech in the form of interviews and folktales, each pair will be too long to train a speech recognition model. Not only that, but when completely done by hand, the transcripts $y_i \in \mathcal{Y}$ may essentialize the utterance by not transcribing mishaps like repeated words. Hence, a new dataset $\tilde\mathcal{D} = \{(\tilde x_j, \tilde y_j)\}$ must be processed where $\tilde y_j$ is the exact transcript of $\tilde x_j$ and all $\tilde x_j$ are singificantly shorter (see Fig. \ref{fig:transcription}).

\begin{figure}[t]
\centering
\includegraphics[width=\columnwidth]{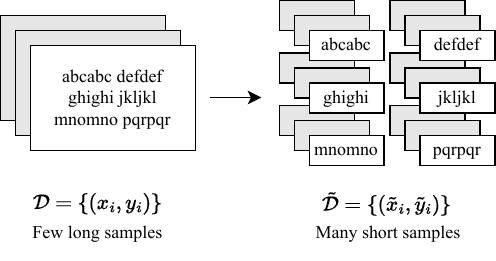}
\caption{Data from the language documentation effort will not be formatted for machine learning tasks and must be processed accordingly.}
\label{fig:dataset}
\end{figure}

The preliminary data for C. Urmi consisted of 8 examples, some as long as 11 minutes in length. This was carefully segmented into roughly 600 samples each no longer than 15 seconds.

\section{Training an ASR Model}

Training an ASR model allows us to transcribe future collected samples more efficiently. Intuitively, the ASR model, which might not be very accurate trained on only on the initial dataset, will output an initial draft of the transcription which human oversight can correct. In the case of C. Urmi, a wav2vec2.0 model fine-tuned on Persian and C. Urmi data achieved a CER of 12.5\% and expidited transcriptions up to $6.3\times$. This model has been made publically available\footnote{\url{https://huggingface.co/mnazari/wav2vec2-assyrian}} to again encourage follow-up work.

\subsection{Deep Learning Over Classical Models}
With the immense quantity of speech data available today, deep learning has outperformed all traditional approaches to ASR \cite{malik2021automatic}. However, training these models can require up to 10,000 hours of labelled data, whereas traditional approaches generally require orders of magnitude less data. Training these traditional models is an extremely intensive process that would be difficult to perform repeatedly in NoLoR. This training data must also be phoneme-aligned, which does not work if simply collecting the unlabelled speech data is challenging enough.

\subsection{Selecting a wav2vec 2.0 Checkpoint}
Thankfully, end-to-end models like wav2vec 2.0 \cite{wav2vec} leverage pre-training to frontload the learning of human speech representations. This means that with very little training data, a wav2vec 2.0 model pretrained on 100,000+ hours of 53 multilingual speech data can be fine-tuned for other tasks \cite{chen2021exploring}, including ASR for other languages.

The wav2vec 2.0 architecture takes in as input raw waveform $\mathbf{x}_i \in \mathcal{X}$. An encoder network $f : \mathcal{X}^u \to \mathcal{Z}$ comprised of convolutional layers outputs a feature representation $\mathbf{z}_i \in \mathcal{Z}$ of the original waveform. Context is encoded into the feature representation using the context network $g : \mathcal{Z}^v \to \mathcal{C}$. This network mixes multiple latent representations $\mathbf{z}_i,\, \mathbf{z}_{i-1},\, \ldots,\, \mathbf{z}_{i-v}$ into a single contextualized representation $\mathbf{c}_i$ using more convolutional layers. These contextualized representations can then be used as inputs into a language model and decoded into text.

\begin{figure}[t]
    \centering
    \includegraphics[width=\columnwidth]{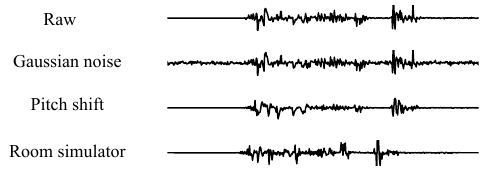}
    \caption{Data Augmentation}
    \label{fig:aug}
\end{figure}

The encoder and context networks are trained by self-learning from an enourmous amount of multilingual data. The initial dataset will not be large enough to fine-tune these layers to the specific phonology of the language of interest. If we don't fine-tune these layers, we run risk of being handicapped by generic contextualized representations. Hence, in the case of C. Urmi, we use a wav2vec 2.0 checkpoint fine-tuned on a phonologically adjacent language, Persian.

\subsection{Language Model}
A character-level tokenizer was used as a language model. The alternate to this would be using a subword tokenizer with a relatively small vocabulary size of 256 or 512. This was attempted but proved to have poor results. Training a subword tokenizer requires sample text to learn the the frequencies of each subword. However, with such a limited amount of text data in existance, such a tokenizer proved to be too specific and unable to identify novel words or verb inflections. This behavior is expected according to Zipf's law.

The loss used to decode the contextualized representations was a connectionist temporal classification (CTC) loss \cite{Graves2012}. CTC loss is necessary for training on data that is not phoneme-aligned.

\subsection{Fine-tuning}
In order to fine-tune from the wav2vec 2.0 checkpoint, the encoder and context networks were frozen during training and the CTC language model was concatenated to the end of the context network for fine-tuning. By feeding the contextualized representations into a dense layer, a probability distribution over possible characters is generated which can be fed into a CTC loss.

Overfitting on the training split of the dataset was a huge issue, as soon discussed. A very careful sweep of hyperparameters had to be done to discover which combination had the best regularizing effects. 

\subsection{Data Augmentation}

As will be the case in many language documentaiton efforts, the number of speakers comprising the initial dataset may be very low. In the case of our dataset, all speech came from a single elderly woman. Machine learning models are interpolators, meaning it is often difficult to generalize their performance on out-of-distribution data \cite{neyshabur2017exploring}.

\begin{table}[t]
    \centering
    \resizebox*{\columnwidth}{!}{
    \begin{tabular}{cccccc}\toprule
        Length  & \multicolumn{2}{c}{Without ASR} & \multicolumn{2}{c}{With ASR} & Speedup \\
              & {\small Time}   & {\small CER (\%)} & {\small Time}  & {\small CER (\%)}  \\ \midrule
        15sec & \phantom{00}3min & 0.0              & 89sec           & 0.0 & 2.0$\times$ \\
        1min  & 25min            & 0.0              & \phantom{0}7min & 0.0 & 3.6$\times$ \\
        3min  & \phantom{0}91min & 0.5              & 17min           & 1.1 & 5.3$\times$ \\
        5min  & 132min           & 0.8              & 21min           & 2.1 & 6.3$\times$ \\ \bottomrule
    \end{tabular}
    }
    \caption{Transcribing 8 unique speech samples with and without ASR at varying lengths.}
    \label{table:results}
\end{table}

In our situation, it means the ASR model will overtrain on this specific speaker's voice by learning that her unique pronunciation habits are instrinsic to C. Urmi as a whole. We use data augmentation to combat this issue, which is a proven technique to improve model robustness and accuracy in low-resource NLP settings \cite{feng2021}. For each example, we apply augmentations (see Fig. \ref{fig:aug}) each at varying degrees. The augmentations were chosen specifically to counteract the speed of the speaker's utterance, the pitch of the speaker's voice, and the clarity of the microphone.

\subsection{Evaluation and Impact}

To evaluate the ASR model, we looked to the diasporic Assyrian population of Armenia and travelled to the southern Assyrian villages near Verin Dvin. Here, we collected several samples of people speaking about the history of the village as well as various jokes and anecdotes. One striking story, in particular, comes from a grandmother who describes in grief how she is discouraged to speak to her children in her language since her daughter had married an Armenian outside of the village. With our work, her voice and story will be documented.

Regarding the efficacy of the ASR model, we clearly observe dramatic improvements to the speed of documentation when assisted by the ASR model (see Table \ref{table:results}). Although the ASR model achieved a test accuracy of 12.5\% CER, the accuracy for certain samples reached as low as 27.5\%. In these examples, the speakers exhibited alternate pronunciations of certain phones (like \textipa{[\textdyoghlig]} for \textipa{/\textObardotlessj/} rather than \textipa{[G]}). This articulation, less common in the C. Urmi dialect \cite{khan2016neo}, had become ubiquitous in this diasporic community. Additionally, when inputting several minutes into the ASR at once, entire sentences were misconstrued which  decreased the accuracy significantly. However, given these challenges, the ASR still provided adequate drafts that proved to be great aids in the transcription process, up to $6.3\times$ faster for longer, more difficult transcriptions.

We noticed that with the ASR model, we tended to assume we were done transcribing early, which meant we allowed mistakes to make it through which we corrected later. We observed this behavior even without the ASR model but to a smaller extent.

\begin{figure}[t]
    \centering
    \includegraphics[width=1\columnwidth]{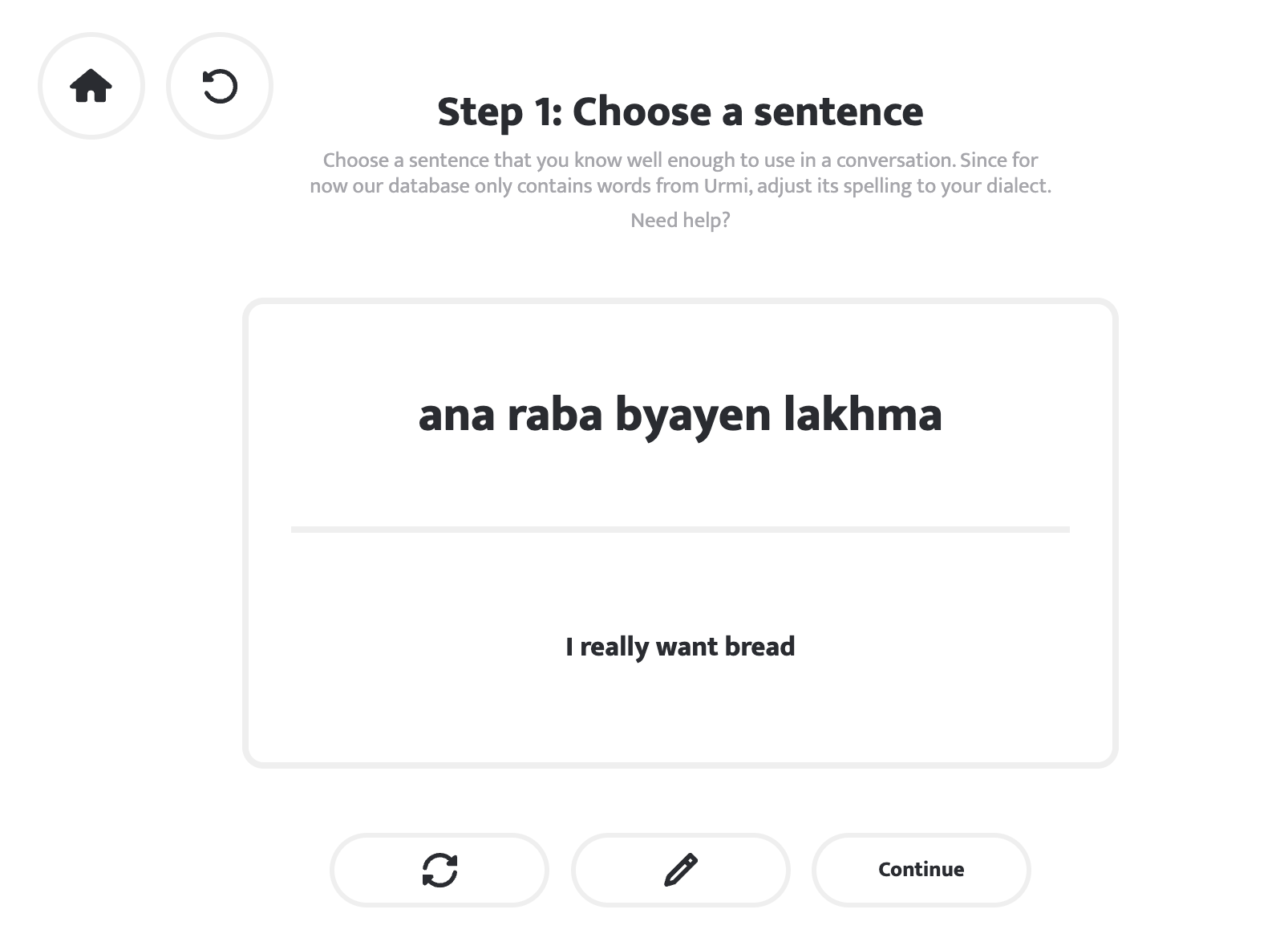}
    \caption{The interface for speakers to find a sentence they would be comfortable reading, which in this case in a simplified spelling.}
    \label{website}
\end{figure}

\section{Expanding the Dataset}
\label{sec:expanding}

To expand this dataset, we collect and transcribe more speech samples as we did in Armenia. Beyond this, however, we developed AssyrianVoices, an online crowdsourcing application for speakers of Neo-Aramaic around the world to donate speech samples.

The application is designed to attentively consider the backgrounds of the users. For example, users are able to specify which dialect they identify with. While the phonemic orthography is useful for transcription and modelling purposes, some characters are not readily legible to non-linguists. In the real world, speakers will use a standard English keyboard and phonetically spell words in a fairly non-standardized way. Some trends appear, such as representing the voiceless postalveolar fricative \textipa{[\v s]} with the sequence \textipa{/sh/} as done in English. A simplification of the phonemic script is hence provided.

\section{Conclusions}

Language documentation is vital to the preservation of endangered languages, but transcribing speech samples is a costly and time-consuming process. Within the low-resource NLP community, there is virtually no discussion about how to facilitate the transcription process.

This paper provides the NoLoR framework for expedited endangered language documentation and demonstrates its efficacy using Neo-Aramaic as a case. By doing so, we produced an ASR model that helped document the oral history of the diasporic Assyrian community of Armenia. We then released our dataset of annotated speech samples and launched a crowdsourcing campaign with the goal of collecting data to further explore the role of AI in language documentation.

\subsection{Next Steps}

There are many future steps that will further improve the efficacy of NoLoR as well as develop it into a more out-of-the-box approach. One such improvement would involve building a model to segment the long-form audio samples automatically in order to shorten the step of building an initial dataset. Another direction is to implement a system where fieldworkers can simply dump their new transcribed samples and have the ASR model periodically fine-tune itself. This way, an engineer does not have to be on staff in order to continuously improve the ASR model. The most immediate work, however, is to continue language documentation efforts with C. Urmi and see how many iterations of NoLoR is necessary to eliminate the need for human intervention entirely.

An area of interest specific to Neo-Aramaic is creating programs to educate members of the community of the importance of documentation efforts like AssyrianVoices and teach them how to use it.

\section{Acknowledgements}
My loving mother, who serves as a living reminder that our language is worth saving, was the inspiration for this work. Such an undertaking would not have been possible if not for Geoffrey Khan and him relentlessly applying time, talent, and adoration to preserving the linguistic and cultural heritage of the Assyrian people.

\bibliography{aaai23}

\end{document}